\documentclass[10pt,twocolumn,letterpaper]{article}

\usepackage{iccv}
\usepackage{times}
\usepackage{epsfig}
\usepackage{graphicx}
\usepackage{amsmath}
\usepackage{amssymb}
\usepackage{multirow}
\usepackage{tabularx}

\usepackage[pagebackref=true,breaklinks=true,letterpaper=true,colorlinks,bookmarks=false]{hyperref}

\iccvfinalcopy % *** Uncomment this line for the final submission

\def\iccvPaperID{350} % *** Enter the ICCV Paper ID here

\begin{document}

%%%%%%%%% TITLE
\title{\vspace{-5mm}DeepContext: Context-Encoding Neural Pathways for \\ 3D Holistic Scene Understanding\vspace{-5mm}}

%\author{Yinda Zhang\\
%Princeton University\\
%% For a paper whose authors are all at the same institution,
%% omit the following lines up until the closing ``}''.
%% Additional authors and addresses can be added with ``\and'',
%% just like the second author.
%% To save space, use either the email address or home page, not both
%\and
%Mingru Bai\\
%Princeton University\\
%\and
%Pushmeet Kohli\\
%OpenAI\\
%\and
%Shram Izadi\\
%Perspective IO\\
%\and
%Jianxiong Xiao\\
%AutoX Inc.
%}

\author{Yinda Zhang$^1$~~~~Mingru Bai$^1$~~~~Pushmeet Kohli$^{2,5}$~~~~Shahram Izadi$^{3,5}$~~~~Jianxiong Xiao$^{1,4}$\\ \vspace{-3mm} \\ $^1$Princeton University~~~~$^2$DeepMind~~
~~$^3$PerceptiveIO~~~~$^4$AutoX~~~~$^5$Microsoft Research }

\maketitle
\vspace{-2mm}

%%%%%%%%% ABSTRACT
\begin{abstract}
3D context has been shown to be extremely important for scene understanding, yet very little research has been done on integrating context information with deep neural network architectures. This paper presents an approach to embed 3D context into the topology of a neural network trained to perform holistic scene understanding.
Given a depth image depicting a 3D scene, 
our network aligns the observed scene with a predefined 3D scene template, 
and then reasons about the existence and location of each object within the scene template.
In doing so,
our model recognizes multiple objects in a single forward pass of a 3D convolutional neural network,
capturing both global scene and local object information simultaneously.
To create training data for this 3D network, we generate partially synthetic depth images which are rendered by replacing real objects with a repository of CAD models of the same object category\footnote{Code and dataset are available at \href{http://deepcontext.cs.princeton.edu}{http://deepcontext.cs.princeton.edu}. Part of this work is done when Yinda Zhang was an intern at Microsoft Research, Jianxiong Xiao was at Princeton University, Pushmeet Kohli and Shahram Izadi were at Microsoft Research.}.
Extensive experiments demonstrate the effectiveness of our algorithm compared to the state of the art.
%The dataset and source code of this project will be made available.
%Source code and data will be available. % on GitHub. %upon acceptance.
% non-deep learning context models.
%\keywords{Scene Understanding, Deep Learning, Context, RGB-D Data}

\end{abstract}

%%%%%%%%% BODY TEXT
\vspace{-1mm}
\section{Introduction}
Understanding indoor scene in 3D space is critically useful in many applications, such as indoor robotics, augmented reality.
To support this task, the goal of this paper is to recognize the category and the 3D location of furniture from a single depth image.

Context has been successfully used to handle this challenging problem in many previous works. Particularly, holistic scene context models, which integrate both the bottom up local evidence and the top down scene context, have achieved superior performance \cite{choi2013understanding,lin2013holistic,liu2014creating,PanoContext,zhaointegrating}.
However, they suffer from a severe drawback that the bottom up and top down stages are run separately.
The bottom up stage using only the local evidence needs to generate a large quantity of noisy hypotheses to ensure a high recall, and the top down inference usually requires combinatorial algorithms, such as belief propagation or MCMC, which are computationally expensive in a noisy solution space. Therefore, the whole combined system can hardly achieve a reasonably optimal solution efficiently and robustly.

%However, they suffer from several drawbacks.
%The bottom up and top down stages are often run separately, such that the context model has to select from a huge number of noisy hypotheses generated using weak local evidence.
%Additionally, the inference of these context models usually require combinatorial algorithms, such as belief propagation or MCMC, which are computationally expensive, unstable, and non-optimal.
Inspired by the success of deep learning, we propose a 3D deep convolutional neural network architecture that jointly leverages local appearance and global scene context efficiently for 3D scene understanding.

\begin{figure}[t]
% \vspace{-2mm}
\includegraphics[width=1\linewidth]{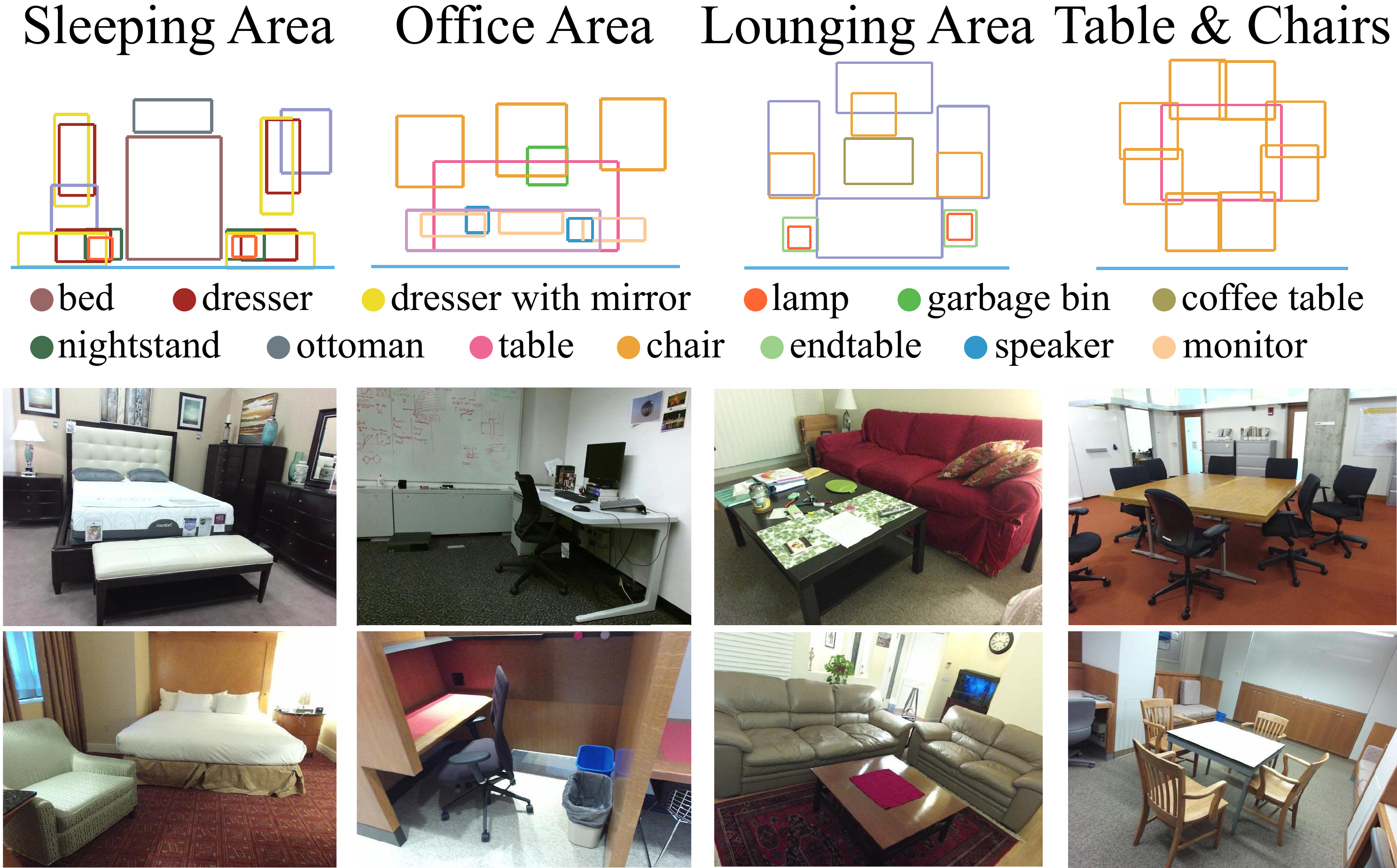}
\vspace{-2mm}
\caption{{\bf Example of canonical scene templates (top view) and the natural images they represent.} We learn four scene templates from SUN-RGBD\cite{SUNRGBD}. Each scene template encodes the canonical layout of a functional area.
%We define a major objects for each of the four scene templates, which are bed, table, sofa, and table from left to right.
}

\label{fig:scene_template_example}
\vspace{-2mm}
\end{figure}

\begin{figure*}[t]
\vspace{-3mm}
\includegraphics[width=1\linewidth]{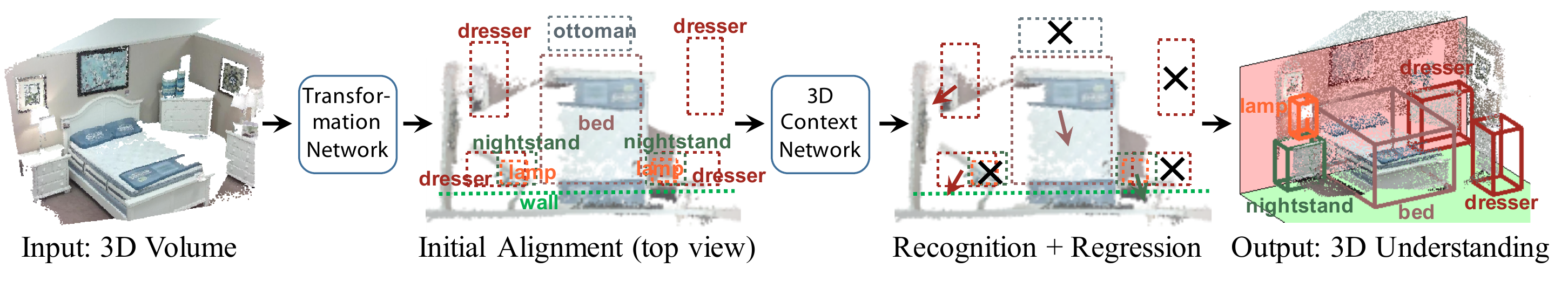}

\vspace{-1mm}
\caption{{\bf Our deep 3D scene understanding pipeline.} Given a 3D volumetric input derived from a depth image, 
we first aligns the scene template with the input data.
Given the initial alignment, our 3D context network estimates the existence of an object and adjusts the object location based on local object features and holistic scene feature,
to produce the final 3D scene understanding result.}
\label{fig:teaser}
\vspace{-4mm}
\end{figure*}

Designing a deep learning architecture to encode context for scene understanding is challenging.
Unlike an object whose location and size can be represented with a fixed number of parameters, a scene could involve unknown number of objects and thus requires variable dimensionality to represent, which is hard to incorporate with convolutional neural network with a fixed architecture.
Also, although holistic scene models allow flexible context, they require common knowledge to manually predefine relationship between objects, e.g. the relative distance between bed and nightstands.
As a result, the model may unnecessarily encode weak context, ignore important context, or measure context in an over simplified way.

To solve these issues, we propose and learn a scene representation encoded in scene templates.
A scene template contains a super set of objects with strong contextual correlation that could possibly appear in a scene with relatively constrained furniture arrangements. 
It allows a prediction of ``not present'' for the involved objects so that a variety of scenes can be represented with a fixed dimensionality.
%It can be represented with a fixed dimension of parameters, such that can be directly deployed in the deep learning architecture.
A scene can be considered as a scene template with a subset of objects activated.
Scene template also learns to only consider objects with strong context, and we argue that context-less objects, such as a chair can be arbitrarily placed, should be detected by a local appearance based object detector.

Each template represents a functional sub-region of an indoor scene, predefined with canonical furniture arrangements and estimated 3D anchor positions of possible objects with respect to the reference frame of the template.
We incorporate these template anchors as priors in the neural architecture by designing a transformation network that aligns the input 3D scene (corresponding to the observed depth image) with the template (i.e. the canonical furniture arrangement in 3D space). 
%Unlike [14] where the transformation used for the alignment of the observation and canonical view is learned in an unsupervised fashion, we define the desired alignment in template coordinates and use supervised training by employing the ground truth alignments available from our training data. 
The aligned 3D scene is then fed into a 3D context neural network that determines the existence and location of each object in the scene template.
This 3D context neural network contains a holistic scene pathway and an object pathway using 3D Region Of Interest (ROI) pooling in order to classify object existence and regress object location respectively. 
Our model learns to leverage both global and local information from two pathways, and can recognize multiple objects in a single forward pass of a 3D neural network.
It is noted that we do not manually define the contextual relationships between objects, but allow the network to automatically learn context in arbitrary format across all objects.

Data is yet another challenging problem for training our network. %Compared to the traditional object detection where an infinite number of training patches could be obtained from a single image, our network takes the whole scene as input and needs to model different variations in the scene layout.
% Hence, our network requires more training data. 
% %SUN-RGBD dataset contains 10,335 scene images, however only 7379 of them are taken from real indoor scene, and the others are from furniture store and classroom which does not contain useful context.
% Empirically, we found data from SUN-RGBD are far from sufficient to train our neural network.
% Therefore, we propose a hybrid data augmentation method to create synthetic depth image.
% We take the scene image from SUN-RGBD, and replace the point cloud within the bounding box of each object with the depth rendered from CAD model.
% Our synthetic data exhibits a variety of local object appearance, while still keep the indoor furniture arrangement and clutter as shown from real scene.
Holistic scene understanding requires the 3D ConvNet to have sufficient model capacity, which needs to be trained with a massive amount of data. %, which is a problem as . 
However, existing RGB-D datasets for scene understanding are all small.
To overcome this limitation, we synthesize training data from existing RGB-D datasets by replacing objects in a scene with those from a repository of CAD models from the same object category, and render them in place to generate partially synthesized depth images.
Our synthetic data exhibits a variety of different local object appearances, while still keeping the indoor furniture arrangements and clutter as shown in the real scenes.
In experiments, we use these synthetic data to pretrain and then finetune our network on a small amount of real data, whereas the same network directly trained on real data can not converge.

The contributions of this paper are mainly three aspects. 1) We propose a scene template representation that enables the use of a deep learning approach for scene understanding and learning context. The scene template only encodes objects with strong context, and provides a fixed dimension of representation for a family of scenes. 2) We propose a 3D context neural network that learns scene context automatically. It leverages both global context and local appearance, and detects all objects in context efficiently in a single forward pass of the network. 3) We propose a hybrid data augmentation method, which generates depth images keeping indoor furniture arrangements from real scenes but containing synthetic objects with different appearance.

\vspace{-3mm}
\paragraph{Related Work}
The role of context has been studied extensively in computer vision~\cite{SimulationScene,choi2010exploiting,choi2012context,choi2012tree,desai2011discriminative,SongChunZhu05,GrammarParsing,heitz2008cascaded,ladicky2010graph,li2011feccm,li2009towards,li2010object,ProbabilisticGraphics,mottaghi2014role,oliva2007role,Context,sudderth2005describing,sudderth2008shared,sudderth2005learning,sudderth2008describing,GrowMind,tu2008auto,tu2005image,AndOrGraph}.
While most existing research is limited to 2D, there are some works on modeling context for total scene understanding from RGB-D images~\cite{RGBDcuboid,lin2013holistic,Brown2016,sudderth2006depth,PanoContext}. In term of methodology, most of such approaches take object detection as the input and incorporate context models during a post-processing. We aim to integrate context more tightly with deep neural network for object detection.

There are some efforts incorporating holistic context model for scene understanding, which is closely related to our work. Scene context is usually manually defined as a unary term on a single object, pairwise term between a pair of objects to satisfy certain functionality \cite{lin2013holistic,yu2011make}, or a more complicated hierarchy architecture \cite{choi2013understanding,liu2014creating,zhaointegrating}. 
The learned context models are usually applied on a large set of object hypotheses generated using local evidence, e.g. line segments \cite{zhaointegrating} or cuboid \cite{lin2013holistic}, by energy minimization.
Therefore high order context might be ignored or infeasible to optimize.
Context can be also represented in a non-parametric way \cite{PanoContext}, which potentially enables high order context but is more computationally expensive to infer during the testing time.
%These context models are flexible, however require heuristically predefine the objects involved and the analytical formulation, e.g. pairwise, such that some potentially useful context, especially in high order, may not be leveraged. 
In contrast, our 3D context network does not require any heuristic intervene on the context and learns context automatically.
We also require no object hypothesis generation, which is essential in making our method more computationally efficient.

%The context is usually applied by minimizing over an energy function over an object hypotheses space containing thousands of proposals, such that these algorithm are usually slow and high order context is hard to be optimized. Comparatively, our scene template only provides a representation of the input and output, and does not require any further heuristic definition of the context and allow network to learn fully automatically. We also require no object hypothesis generation, but only a single pass of the whole scene to detect all the object, which is efficient.

% yinda: one contribution: We process the whole scene more efficiently thanks to the DNN, and the context model learned is more powerful. As a result, our system still works with fixed proposals, i.e. scene template, which does not require proposal generation.

%The context within scene is formulated by a variety of pairwise constraints encoding relative spatial relations

Deep learning has been applied to 3D data, but most of these works focus on modeling objects~\cite{3DShapeNets} and object detection~\cite{maturana2015voxnet,DeepSlidingShapes}. 
Recently, some successes have been made on applying deep learning for inverse graphics \cite{kulkarni2015picture,kulkarni2015deep}.
Our approach goes one step further to embrace the full complexity of real-world scenes to perform holistic scene understanding. 
Related to our transformation network, Spatial Transformation Networks \cite{jaderberg2015spatial} can learn the transformation of an input data to a canonical alignment in an unsupervised fashion. 
However, unlike MNIST digits (which were considered in \cite{jaderberg2015spatial}) or an individual object where an alignment to a canonical viewpoint is quite natural, it is not clear what transforms are needed to reach a canonical configuration for a 3D scene. We define the desired alignment in template coordinates and use supervised training by employing the ground truth alignments available from our training data. 

While many works have considered rendering synthetic data for training (a.k.a, graphics for vision, or synthesis for analysis),
these efforts mostly focus on object rendering, either in color \cite{li2015joint,Su_2015_ICCV} or depth \cite{SlidingShapes}. 
There is also work rendering synthetic data from CAD model of complicated scenes for scene understanding \cite{handa2015scenenet,zhang2016deep}. However, the generated depth is overly clean, and the scene layouts generated by either by algorithm or human artists are not guaranteed to be correct. 
In contrast, we utilize both the CAD models and real depth maps to generate more natural data with appropriate context and real-world clutter. 

%\cite{lin2013holistic,yu2011make} flexible, pairwise, unpredicted, heuristically defined. We fully auto

%\cite{zhaointegrating,choi2013understanding} similar idea, but require hypothesis generation and optimization which may not be optimal

% draft:
% goal
% context approach
% problem: bottom up, top down, combinatory, belief propagation, slow
% approach: deep learning encode context
% challenge:
% template design
% data augmentation

% \vspace{-1mm}
\section{Algorithm Overview}
\vspace{-1mm}

Our approach works by first automatically constructing a set of scene templates from the training data (see Section~\ref{sec:learn_template}). 
Rather than a holistic model for everything in the scene, each scene template only represents objects with context in a sub-area of a scene performing particular functionality. 
Each template defines a distribution of possible layouts of one or more instances of different object categories in a fixed dimensionality.
%We train a 3D neural network for each scene template to perform 3D object detection with a single pass of the data.

Given a depth map of a scene as input\footnote{Note that while all the figures in the paper contain color, our system relies only on depth as input without using any color information.}, we convert it into a 3D volumetric representation of the scene and feed it into the neural network.
The neural network first infers the scene template that is suitable to represent the scene, or leaves it to a local appearance based object detector if none of the predefined scene templates is satisfied. 
If a scene template is chosen, the transformation network estimates the rotation and translation that aligns the scene to the inferred scene template.
With this initial alignment, the 3D context  network extracts both the global scene feature encoding scene context and the local object features pooled for each anchor object defined in the template, as shown in Fig.\ref{fig:teaser}.
These features are concatenated together to predict the existence of each anchor object in the template and an offset to adjust its bounding box for a better object fit.
The final result is an understanding of the scene with a 3D location and category for each object in the scene, as well as room layout elements including wall, floor, and ceiling, which are represented as objects in the network.

\begin{figure}[t]
\vspace{-2mm}
\begin{center}
\includegraphics[width=1\linewidth]{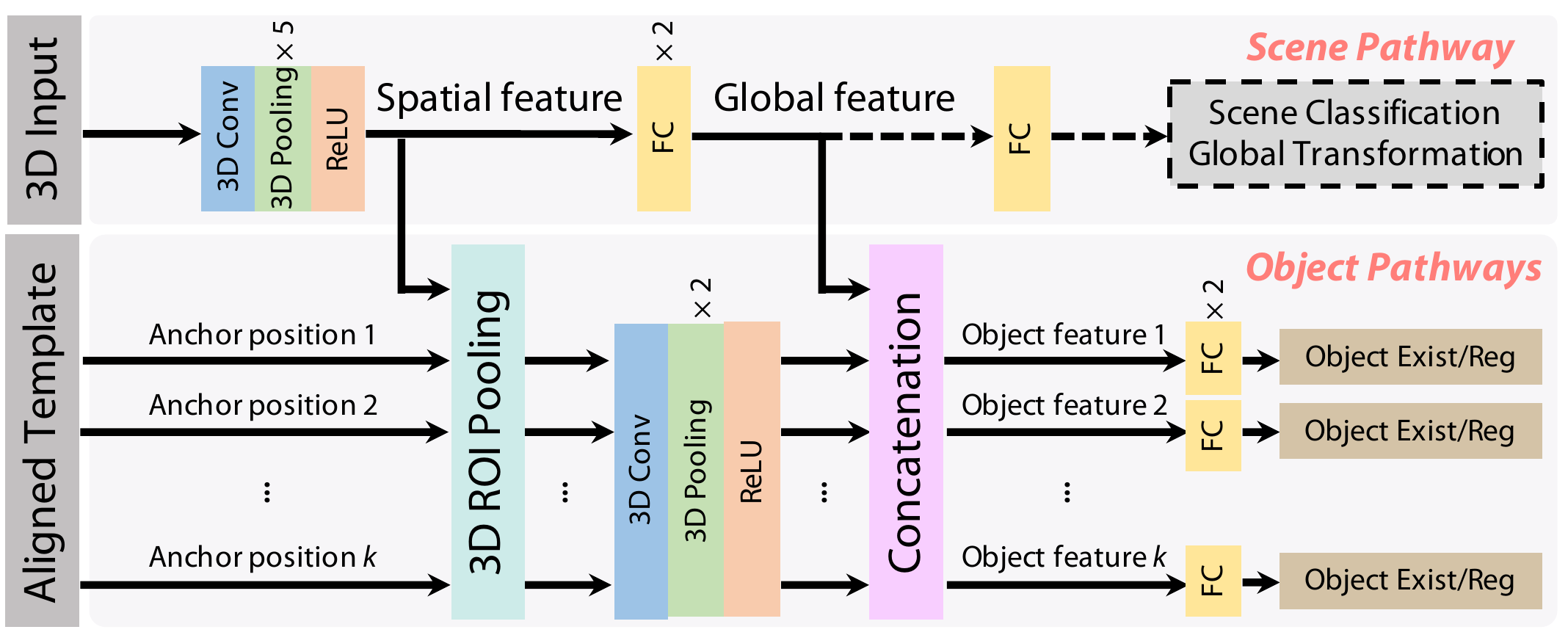}
\end{center}

\vspace{-2mm}
\caption{{\bf 3D context network.}
The network consists of two pathways. The scene pathway takes the whole scene as input and extracts spatial feature and global feature. The object pathway pools local feature from the spatial feature. The network learns to use both the local and global features to perform object detection, including wall, ceiling, and floor.
}
\vspace{-2mm}
\label{fig:network_structure}
\end{figure}

\vspace{-2mm}
\section{Learning Scene Template}
\vspace{-1mm}
Objects with context in a functional area are usually at relatively fixed locations. 
For example, a sleeping area is usually composed of a bed with one or two nightstands on the side, with optional lamps on the top. 
Object detection is likely to succeed by searching around these canonical locations.
We learn the categories of object instances and their canonical sizes and locations in the template, from the training data.
Examples of each template can be seen in Fig. \ref{fig:scene_template_example}. 

% \begin{figure}[t]
% \vspace{-2mm}

% \includegraphics[width=1\linewidth]{figures/data_assignment_new.pdf}

% \vspace{-4mm}

% \caption{{\bf Converting annotation into the scene template.} 
% (a) After aligned with scene template via pose of major object, a bipartite matching then assigns objects from SUNRGBD to anchor objects in template. (b) Example of scene template ground truth. Multiple instances of a same object category are marked with the index of the same color.
% %Please refer to Section \ref{sec:contextNet} for more details.
% }
% \label{fig:ground_truth}
% \vspace{-5mm}
% \end{figure}

\subsection{Data-driven Template Definition}
\vspace{-1mm}
\label{sec:learn_template}
We learn to create scene templates using the SUN-RGBD dataset consisting of 10,335 RGB-D images with 3D object bounding box annotations.
These RGB-D images are mostly captured from household environments with strong context.
As a first experiment of combining 3D deep learning with context, we choose four scene templates: sleeping area, office area, lounging area, and table \& chair set, because they represent commonly seen indoor environments with relatively larger numbers of images provided in SUN-RGBD.
Our approach can be extended to other functional areas given sufficient training data.
For SUN-RGBD, $75\%$ of the images from household scene categories can be described, fully or partially, using these four scene templates. 

Our goal is to learn layouts of the scene, such that each template summarizes the bounding box location and category of all objects appearing in the training set.
To enable the learning of the template, we select the images that contain a single functional area, and label them with the scene type they belong to.
Other images containing arbitrary objects or multiple scene templates are not used in learning scene templates. 
The ground truth scene categories are used not only for learning the aforementioned templates, but also for learning the scene template classification, the transformation networks, and the 3D context networks in the following sections.

To obtain the anchor positions (i.e. common locations) for each object type in a template, we take all 3D scenes belonging to this scene template and align them with respect to the center and orientation of a major object\footnote{We manually choose bed for sleeping area, desk for office area, sofa for lounging area, and table for table\&chair set as the major objects.}. 
After that, we run $k$-means clustering for each object type and use the top $k$ cluster centroids as the anchor positions and size, where $k$ is user-defined.
We also include room layout elements, including wall, floor, ceiling, which are all represented as regular objects with predefined thickness.
% In this way, we learn 3D scene templates from the training data.
Each scene template has tens of object anchors in total for various object categories (Fig.~\ref{fig:scene_template_example}).
% Note that multiple instances of a category could appear with overlapping.

\subsection{Generating Template-Based Ground Truth}
\vspace{-1mm}
To train a 3D context network using scene templates, 
%we need to have ground truth aligned with the template.
%Therefore, 
we need to convert the original ground truth data from SUN RGB-D dataset to a template representation.
Specifically, we need to associate each annotated object in the original ground truth with one of the objects defined in the scene template.
% Fig. \ref{fig:ground_truth}(a) illustrates the following procedure by an example.
% The procedure is as the follows.
Similar to above, we first align the training images with their corresponding scene templates using the center and rotation of the major object.
%Given a training image, we find the major object, translate its center to the origin of the coordinate system, and rotate its orientation to align with that of the template. 
For the rest of the objects, we run a bipartite matching between the dataset annotation and the template anchors, using the difference of center location and size as the distance, while ensuring that the objects of the same category are matched.
% Fig. \ref{fig:ground_truth}(b) shows some examples. 

% \vspace{-1mm}
\section{3D Scene Parsing Network}
\vspace{-1mm}
Given a depth image as input, we first convert it into a 3D volumetric representation, using the Truncated Signed Distance Function (TSDF) \cite{DeepSlidingShapes,KinectFusion}. 
We use a $128\times 128 \times 64$ grid for the TSDF to include a whole scene, with a voxel unit size of 0.05 meters and a truncation value of 0.15 meters.
This TSDF representation is fed into the 3D neural network such that the model runs naturally in 3D space and directly produces output in 3D.

\subsection{Scene Template Classification Network}
\vspace{-1mm}
\label{sec:template_classification}
We first train a neural network to estimate the scene template category for the input scene (Fig.~\ref{fig:network_structure}, Scene pathway).
The TSDF representation of the input scene is firstly fed into 3 layers of 3D convolution + 3D pooling + ReLU, and converted to a spatial feature map. After passing through two fully connected layers, the 3D spatial feature is converted to a global feature vector that encodes the information from the whole scene. The global feature is used for scene template classification with a classic softmax layer.
During testing, we choose the scene template with the highest score for the input scene if the confidence is high enough ($>0.95$).
Otherwise, we do not run our method because none of the scene templates fits the input scene. Such scenes are passed to a local appearance based object detector for object detection.
In practice, the four scene templates can match with more than half of the images in the SUN-RGBD dataset captured from various of indoor environments.
% Accurate 3D understanding on this portion of data using context achieves in significant improvement upon the state of the art methods.

\subsection{Transformation Network}
\begin{figure}[t]
\vspace{-2mm}
\center
\includegraphics[width=\linewidth]{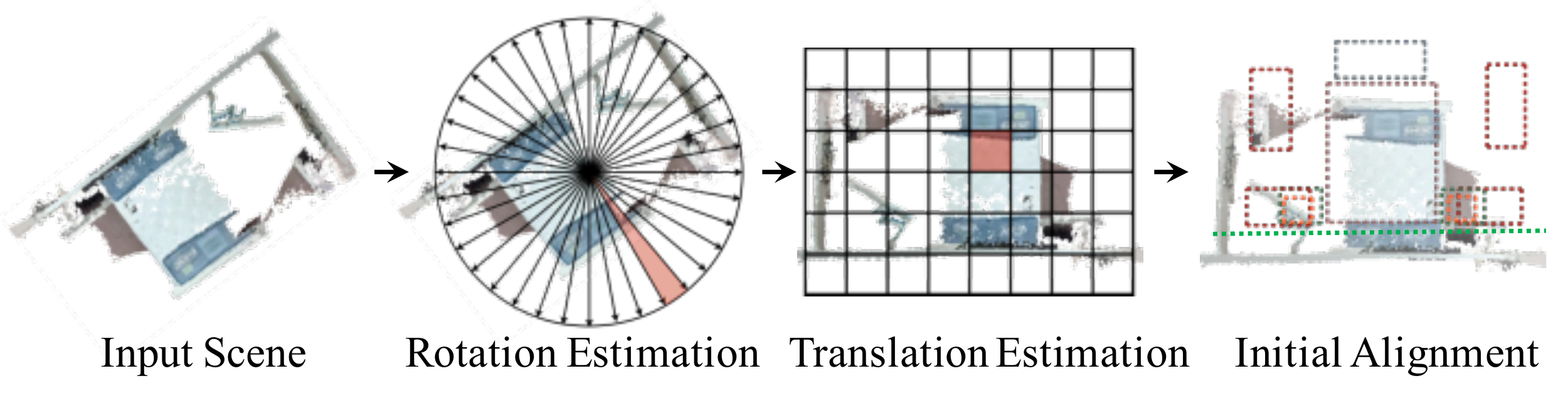}

\caption{{\bf Transformation estimation.} 
Our transformation network first produces global rotation and then translation to align the input scene with its scene template in 3D space.
Both the rotation and translation are estimated as classification problems.
}
\label{fig:transformation}
\vspace{-6mm}
\end{figure}

\vspace{-1mm}
Given the scene template category, our method estimates a global transformation consisting of a 3D rotation and translation that aligns the point cloud of the input scene to the target predefined scene-template (Fig.~\ref{fig:transformation}). 
%This transformation consisting of a 3D rotation and translation aligns the point cloud of the input scene with the 3D scene template as closely as possible. 
This is essentially a transformation that aligns the major object in the input scene with that from the scene template.
%, i.e. rotate the input scene to the upward direction as shown in Fig.~\ref{fig:scene_template_example} and translate the center of major objects coincident.
% Our goal is to transform the input scene, such that it faces towards the same direction with the scene template with the center of the major object coincides with the template.
This makes the result of this stage invariant to rotations in the input, and the wall and bounding box of objects are globally aligned to three main directions.
The next part of our architecture, the 3D context network, relies on this alignment to obtain the object orientation and the location to pool feature based on 3D object anchor locations from the scene template.

%Since we have the ground truth, we directly train network to estimate transformation.
We first estimate the rotation.
% rotation correction enables better translation predictions. 
We assume that the gravity direction is given, e.g. from an accelerometer. In our case, this gravity direction is provided by the SUN RGB-D dataset used in our experiments. 
Therefore, we only need to estimate the yaw, which rotates the input point cloud in horizontal plane to the scene template viewpoint shown in Fig.\ref{fig:scene_template_example}.
We divide the 360-degree range of rotation into 36 bins and cast this problem into a classification task (Fig.~\ref{fig:transformation}).
We train a 3D ConvNet using the same architecture as the scene template classification network introduced in Sec.~\ref{sec:template_classification} except generating a 36 channel output for softmax.
During training, we align each training input scene to the center of the point cloud and add noise for rotations (+/- 10 degrees) and translations ($1/6$ of the range of the point cloud). 

For translation, we apply the same network architecture to identify the translation after applying the predicted rotation.
The goal is to predict the 3D offset between the centers of the major objects of the input point cloud and its corresponding scene template. %from the center of the input point cloud to the center of the main object of the scene template,
%so that in the ideal case, all the scenes from the same template category will be able to align to each other with respect to the center of the major object. 
To achieve this goal, we discretize the 3D translation space into a grid of $0.5\textrm{m}^3$ resolution with dimensions of $[-2.5,2.5]\times[-2.5,2.5]\times[-1.5,1]$,
and formulate this task again as a 726-way classification problem (Fig.~\ref{fig:transformation}). 
We tried direct regression with various loss functions, but it did not work as well as classification.
We also tried an ICP-based approach, however it could not produce good results.

\subsection{3D Context Network}
\vspace{-1mm}
% After aligning the scene template with the input scene, 
We now describe the context neural network for indoor scene parsing using scene templates.
For each scene template defined in the previous section, a separate prediction network is trained.
As shown in Fig.~\ref{fig:network_structure}, the network has two pathways.
The global scene pathway, given a 3D volumetric input in a coordinate
system that is aligned with the template, produces both a spatial feature that preserves the spatial structure in the input data and a global feature for the whole scene.
For the object pathway, we take the spatial feature map from the scene pathway as input, and pool the local 3D Region Of Interest (ROI) based on the 3D scene template for the specific object.
The 3D ROI pooling is a max pooling at $6\times6\times6$ resolution, inspired by the 2D ROI pooling from \cite{girshick2015fast}.
The 3D pooled features are then passed through 2 layers of 3D convolution + 3D pooling + ReLU, and then concatenated with the global feature vector from the scene pathway.
After two more fully connected layers, the network predicts the existence of the object (a binary classification task) as well as the offset of the 3D object bounding box (3D location and size) related to the anchor locations learned in Sec.~\ref{sec:learn_template} (a regression task using L1-smooth loss \cite{DeepSlidingShapes}). 
Including the global scene feature vector in the object feature vector provides holistic context information to help identify if the object exists and its location.
% During testing, we threshold the existence scores at $0.2$ to obtain the final scene understanding results.

\subsection{Training Schema}
\vspace{-1mm}
Our 3D scene parsing network contains a series of components with a large number of parameters.
We perform careful training strategy to avoid bad local optima.
We first train the scene pathway alone to perform a 4-way scene classification task. 
After this training converges, we finetune the classification network to estimate the transformation for each individual scene template.
%using only the data from that specific scene template.
An alternative approach is to jointly train a network for classification and transformation, however this does not perform well in practice.
The object pathway is then enabled, and the two pathways are jointly finetuned to perform object detection.
We found that this form of pretraining, from easy to hard task, is crucial in our experiments.
Otherwise, training the four networks independently from scratch cannot produce meaningful models.

\begin{figure}[t]
% \vspace{-2mm}
\center
\includegraphics[width=\linewidth]{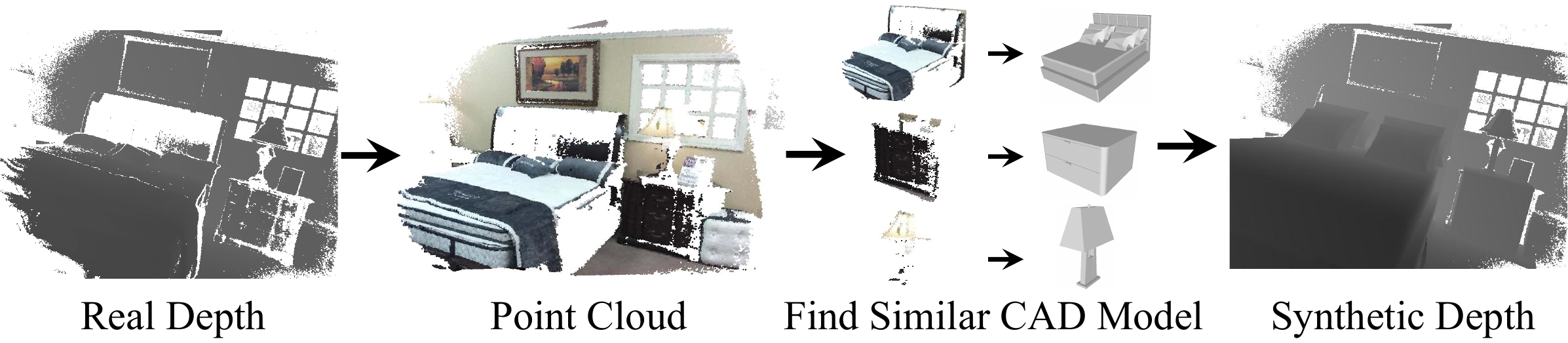}

\caption{{\bf Hybrid data synthesis.} 
We first search for similar CAD model for each object. Then, we randomly choose models from good matches, and replace the points in annotated bounding box with the rendered CAD model.
}
\label{fig:synthetic_data}
\vspace{-4mm}
\end{figure}

\vspace{-1mm}
\section{Synthesizing Hybrid Data for Pre-training}
\vspace{-1mm}
\label{sec:hybrid}
In contrast to existing deep architectures for 3D~\cite{DeepSlidingShapes,3DShapeNets}, our model takes the whole scene with multiple objects as input. As such, during training, it needs to model the different variations in the scene layout. 
We found the RGB-D images from the existing SUN RGB-D \cite{SUNRGBD} dataset are far from sufficient. Furthermore, capturing and annotating RGB-D images on the scale of ImageNet \cite{deng2009imagenet} was impractical. 
To overcome the data deficiency problem, we increase the size of the training data by replacing the annotated objects from SUN RGB-D with CAD models of same category from ShapeNetCore dataset \cite{ShapeNetTechReport} (Fig.~\ref{fig:synthetic_data}). This allows us to generate context-valid scenes, as the context still comes from a real environment, while changing the shapes of the objects. By replacing the annotated objects while keeping the full complexity of the areas outside the annotated bounding boxes, we could generate more realistic hybrid data partially maintaining sensor noise. This is in contrast to images generated from purely synthetic models which do not contain clutter caused by the presence of small objects.

To search for similar CAD models for annotated objects in RGB-D images, we need to define the distance between a CAD model $\mathcal{M}$, and the 3D point cloud $\mathcal{P}$ representing the object. 
In order to get a symmetric definition, we first put the model in the annotated 3D box, scale it to fit, render $\mathcal{M}$ with the camera parameter of the depth image, and convert the rendered depth image to a point cloud $\mathcal{V}$. 
This is to mimic the partial view due to self occlusion. 
Then, we define the distance between $\mathcal{P}$ and $\mathcal{S}$ as:
\vspace{-2mm}\begin{equation*}
D(\mathcal{P},\mathcal{S}) = \frac{1}{\left|\mathcal{P}\right|}\sum_{p\in\mathcal{P}}\Big(\min_{q\in \mathcal{V}}d(p,q)\Big) + \frac{1}{\left|\mathcal{V}\right|}\sum_{p\in\mathcal{V}}\Big(\min_{q\in \mathcal{P}}d(p,q)\Big),
\label{eq:matching}
\vspace{-2mm}
\end{equation*}
where $d(p,q)$ is the distance between two 3D points $p$ and $q$. After acquiring a short list of similar CAD models for each object, we randomly choose one and render the depth image with the original annotation as training data.

We generate a hybrid training set that is 1,000 times bigger than the original RGB-D training set.
For both of the pathways in the 3D context network,
we have to train the models on this large hybrid dataset first, 
followed by finetuning on the real depth maps.
Otherwise, the training cannot converge.

% \begin{table*}[t]
% \center
% \vspace{-2mm}
% {
% \scriptsize
% Rotation Estimation Accuracy~~~~~~~~~~~~~~~~~~~~~~~~~~~~~~~~~~~~~~~~~~~~~~~~~~~~~~~~~~~~~~~~Translation Error (in meters)
% }
% \includegraphics[width=0.95\linewidth]{figures/align_accuracy_cvpr.pdf}
% \vspace{-2mm}
% \caption{{\bf Evaluation of the transformation networks.} 
% Our transformation network outperforms direct point cloud matching in both accuracy of rotation and translation. The proposed testing schema ([Avg.]) consistently works better than a direct testing with single forward ([Single]).
% Please refer to Section~\ref{sec:transformation_alignment} for details.
% }
% \label{fig:alignment_accuracy}
% \vspace{-3mm}
% \end{table*}

% \begin{figure}[t]
% \includegraphics[width=\linewidth]{figures/align_vis.pdf}
% \vspace{-6mm}
% \caption{{\bf Alignment results.} 
% For each scene category,
% we show two successful alignment results and one failure case (in the dashed box). Below each image, we show the point cloud overlaid with the ground truth in original camera coordinates, followed by the aligned result according to the rotation and translation estimated by our network. The red cross marks the origin of the new coordinates, which is expected to be perfect if locates at the center of the major object of each scene.
% }
% \label{fig:align_result}
% \vspace{-4mm}
% \end{figure}

% \vspace{-1mm}
\section{Experiments}
\vspace{-1mm}

We use the SUN RGB-D dataset \cite{SUNRGBD} because they provide high quality 3D bounding box annotations of objects. 
As described in Section \ref{sec:learn_template}, 
we manually select images that can be perfectly represented by one of the scene templates, and choose 1,863 RGB-D images from SUN RGB-D.
%These images are ideal for the training stage to learn scene template and train 3D scene parsing network, and the testing stage to evaluate
%We use 1,502 depth images for training and the remaining 361 depth images for testing.
%We learn scene template and train 3D scene parsing neural network using the training set, and evaluate the performance of scene template classification and transformation estimation on the testing set.
We use 1,502 depth images to learn scene templates and train the 3D scene parsing network, and the remaining 361 images for testing.
We also evaluate our model for object detection on a testing set containing images that cannot be perfectly represented, e.g. containing arbitrary objects or multiple scene templates, to demonstrate that our scene templates have a good generalization capability and a high impact on real scenes in the wild.

% Other evaluation will be shown below.

%We implemented our algorithm using the 3D ConvNets framework Marvin \cite{Marvin}.
Our model uses the half data type, which represents a floating point number by 2 bytes, to reduce the memory usage. We train the model with a mini-batch of 24 depth images requiring 10GB, which nearly fills a typical 12GB GPU.
However, this mini-batch size was too small to obtain reliable gradients for optimization.
Therefore, we accumulate the gradients over four iterations of forward and backward without weight update, and only update the weights once afterwards.
Using this approach, the effective mini-batch size is $24\times 4$ or $96$.

\begin{table*}[t]
\vspace{-2mm}
\centering
\scalebox{0.9}{
\begin{tabular}{l|cccccccccccc}
\hline 
 & \multirow{2}{*}{bed} & night- & \multirow{2}{*}{dresser} & coffee & mirror & end & \multirow{2}{*}{lamp} & \multirow{2}{*}{monitor} & \multirow{2}{*}{ottoman} & \multirow{2}{*}{sofa} & \multirow{2}{*}{chair} & \multirow{2}{*}{table}\tabularnewline
 &  & stand &  & table & dresser & table &  &  &  &  &  & \tabularnewline
\hline 
COG \cite{Brown2016} & 79.8 & 48.1 & 1.70 & - & - & - & - & - & - & 55.8 & 72.9 & 58.4\tabularnewline
DSS \cite{DeepSlidingShapes} & 90.3 & 52.3 & 7.60 & 52.7 & 4.40 & 13.3 & 40.2 & 15.0 & 23.7 & 71.3 & 79.1 & 75.2\tabularnewline
Ours & 89.4 & 63.3 & 19.7 & 40.5 & 16.8 & 27.9 & 41.6 & 18.2 & 13.3 & 50.3 & 44.5 & 65.9\tabularnewline
Ours + DSS & \bf 91.8 & \bf 66.7 & \bf 23.4 & 50.1 & \bf 10.0 & \bf 35.3 & \bf 53.6 & \bf 23.2 & \bf 31.5 & 62.8 & \bf 80.2 & \bf 77.4\tabularnewline
\hline 
GT Align & 92.4 & 64.4 & 19.7 & 49.3 & 23.4 & 25.0 & 31.4 & 16.0 & 15.8 & 63.6 & 46.1 & 70.4\tabularnewline
GT Align+Scene & 94.1 & 66.3 & 19.4 & 48.9 & 23.4 & 21.7 & 31.4 & 16.1 & 15.8 & 74.6 & 50.2 & 74.0\tabularnewline
\hline 
DSS, Full & 75.7 & 30.0 & 7.14 & 19.5 & 0.64 & 11.7 & 20.9 & 1.80 & 8.49 & 51.7 & 52.9 & 41.1\tabularnewline
Ours, Full & \bf 75.8 & \bf 44.1 & \bf 15.7 & \bf 25.8 & \bf 4.99 & \bf 12.4 & \bf 22.4 & \bf 3.47 & \bf 10.7 & 49.0 & \bf 53.2 & 30.5\tabularnewline
\hline 
\end{tabular}
}
\vspace{1mm}
\caption{{\bf Average precision for 3D object detection.} 
We (row 3) achieve comparable performance with DSS \cite{DeepSlidingShapes} (row 2). Combining two methods (row 4) achieves significantly better performance, which shows our model learns context complementary to local appearance. Our model can further achieve better performance with better alignment and scene classification. The last row shows our superior performance on extended testing set where images might not be perfectly represented by any single scene template.}
\label{fig:average_precision}
\vspace{-3mm}
\end{table*}

\subsection{3D object detection.}
\begin{figure*}
\center
\vspace{-5mm}

\includegraphics[width=0.97\linewidth]{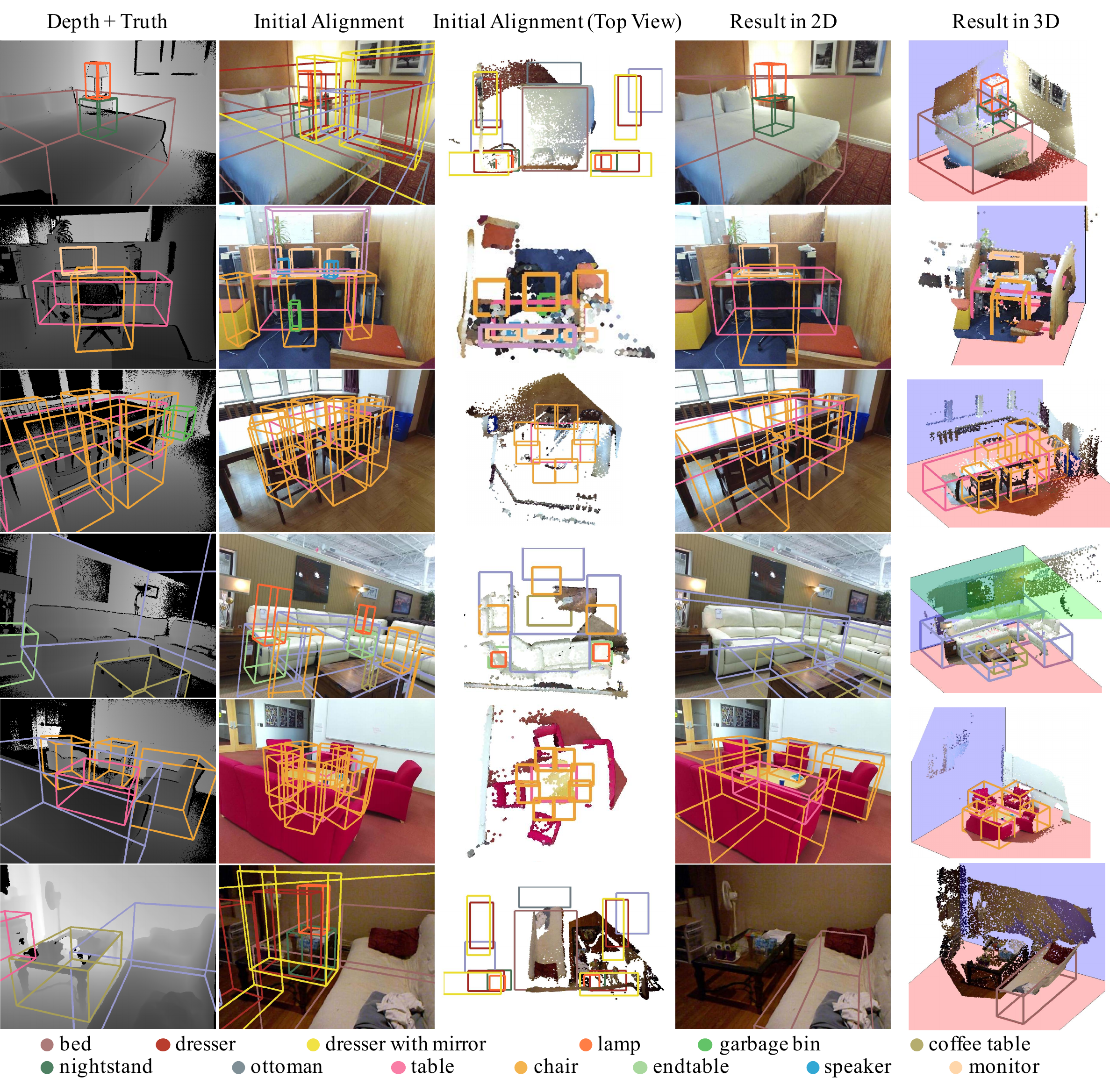}

\vspace{-2mm}
\caption{{\bf Visualization of the qualitative results on the testset.} 
}
\vspace{-3mm}
\label{fig:visualize_result}
\end{figure*}

\vspace{-1mm}
Our model recognizes major objects in a scene, which can be evaluated by 3D object detection.
Qualitative parsing results are shown in Fig. \ref{fig:visualize_result}.
Our model finds most of the objects correctly and produces decent scene parsing results for challenging cases, e.g. heavy occlusion and missing depth. 
3D context enables long range regression when initial alignment is far from correct, as shown in the 5th row. 
The last row shows a failure case,
%For the second to last row, 
%the scene is a hotel room with both the sleeping area and office area visible from the same image.
 where our model recognizes it as a sleeping area misled by the futon with blankets.
 Therefore, our model overlooks the coffee table, but still predicts the wall and floor correctly and find a proper place to sleep.
%For the last row,

% The last row shows a failure
% case. Although it is a lounging area, the futon with blankets
% and clothes makes our system consider it as a sleeping area.
% Therefore, our method recognizes the futon as a bed and
% also predicts the wall and floor correctly, while ignoring the
% coffee table.

Table~\ref{fig:average_precision} shows quantitative comparison to the local appearance based 3D object detector Deep Sliding Shape (DSS) \cite{DeepSlidingShapes}
and also the cloud of gradient feature based context model from Ren \etal (COG) \cite{Brown2016}.
Our average precision (3rd row) is comparable to state-of-the-art, but only takes about 0.5 seconds to process an image for all object categories, which is about 40 times faster than DSS which takes 20 seconds per image.

\begin{figure}[t]
\vspace{-2mm}
\center
\includegraphics[width=\linewidth]{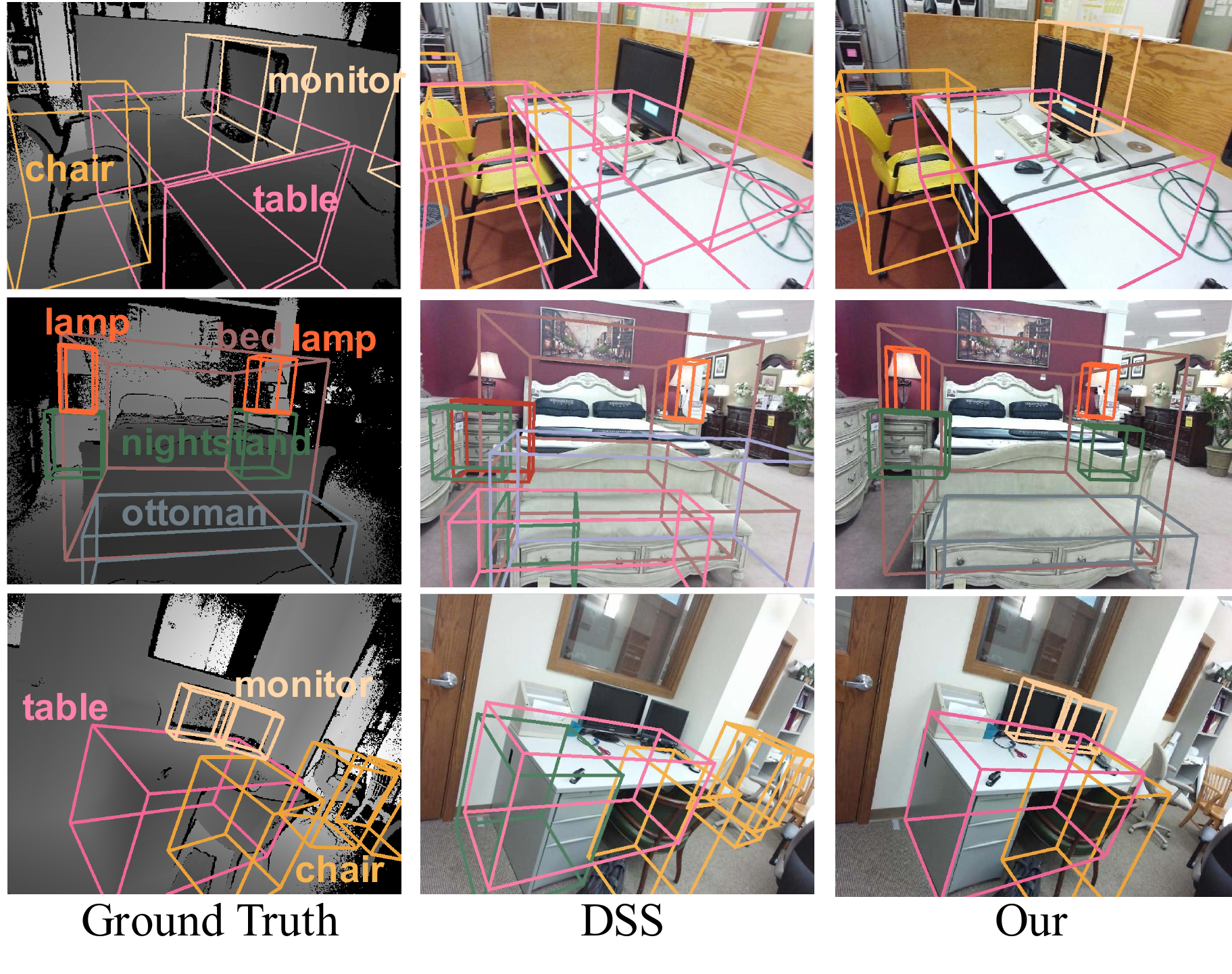}
\vspace{-6mm}
\caption{{\bf Comparison between our context model and the local object detector DSS \cite{DeepSlidingShapes}.} Our context model works well for objects with missing depth (monitors in 1st, 3rd row), heavy occlusion (nightstand in 2nd row), and prevents detections with wrong arrangement (wrong table and nightstand in DSS result). 
}
\label{fig:comparison}
\vspace{-2mm}
\end{figure}

\vspace{-3mm}
\paragraph{Context complements local evidence.}
Fig. \ref{fig:comparison} shows some qualitative comparisons between our context model and the local object detector DSS \cite{DeepSlidingShapes}. We can see that our context model works significantly better in detecting objects with missing depth (the monitor in 1st and 3rd examples) and heavy occlusion (the nightstand in 2nd example). 3D context also helps to remove objects in incorrect arrangements, such as the table on top of another table, and the nightstand at the tail of the bed or in office, as shown in the result of DSS. Comparatively, DSS works better for objects that are not constrained, e.g. chairs on the right of 3rd example.

We integrate the result from DSS and our context model.
The combined result achieves significantly better performance than each of the models individually, increasing the mean average precision from the $43.76\%$ for DSS stand-alone to $50.50\%$.
This significant improvement demonstrates that our context model provides complementary information with a local appearance based object detector.

\begin{figure}[t]
%\vspace{-2mm}
\center
\includegraphics[width=1.0\linewidth]{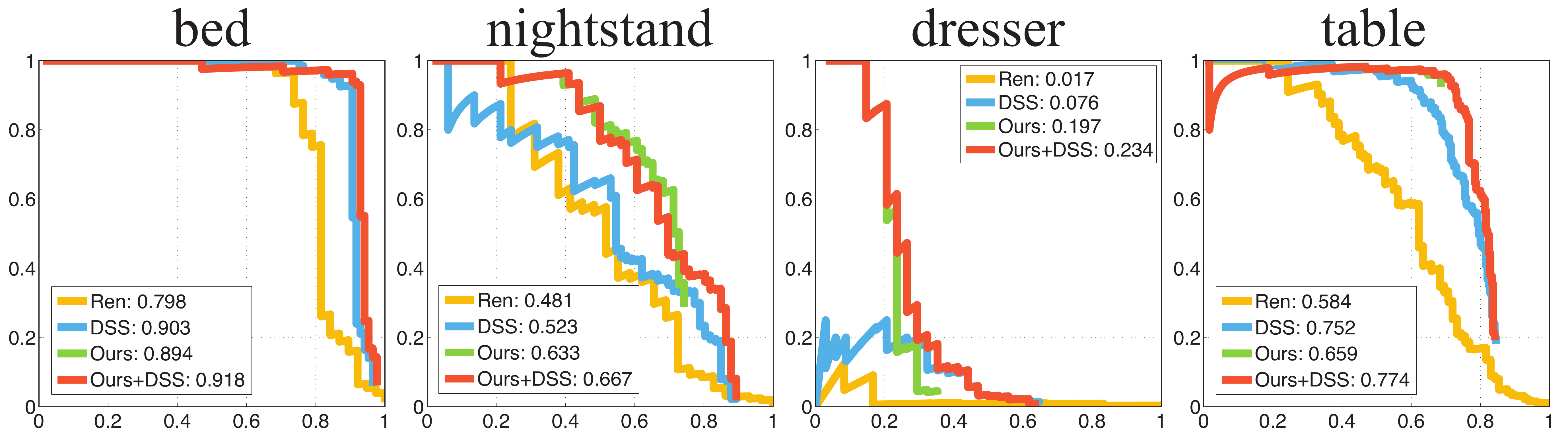}
\vspace{-2mm}
\caption{{\bf Precision recall curves for some object categories.} 
We compare our algorithm with the 3D object detector DSS \cite{DeepSlidingShapes}
and the cloud of gradient feature based context model Ren \etal \cite{Brown2016}.
%We choose a few representative categories that we perform better and worse.
}
\vspace{-4mm}
\label{fig:pr_curve_object}
\end{figure}

Fig.~\ref{fig:pr_curve_object} shows the Precision-Recall (PR) curves for some of the object categories.
We can clearly see that our (green) recalls are not as high as DSS (blue) that runs in a sliding window fashion to exhaustively cover the search space. 
This is because our model only detects objects within the context.
%of the scene templates that have only tens of anchor objects.
% Instead of evaluating object recognition in 2000 boxes as in DSS, we only evaluate tens of boxes defined in the scene template. 
However, our algorithm maintains a very high precision, which applies to a broader range of working situations, with slightly lower recall.
%our algorithm maintains a very high precision and still achieves a reasonable recall even with slightly lower AP than DSS. 
Nevertheless, combining the result of our method and DSS (red) obtains the best performance in terms of both precision and recall.
%Also, our method is nearly 50 times of faster than DSS which takes about 20 seconds to process an image.

\vspace{-3mm}
\paragraph{Generalization to imperfect scene template images. }
Our method can work not only on perfect scene template images, but also images in the wild.
Thanks to the template classification and alignment component, our method can find the right place in the input scene to apply the context model.
To evaluate, we randomly pick 2,000 images that are not used for training from the SUN-RGBD dataset.
This uniformly sampled testing set reflects the scene distribution from the dataset, and contains many images that cannot be perfectly represented by any of the scene templates.
We test DSS on this test set and achieve $26.80\%$ mAP (Table~\ref{fig:average_precision}, the 2nd last row), which is similar to the performance reported in \cite{DeepSlidingShapes}. 
We further run our method on testing images with the template classification confidence higher than 0.95, which ends up choosing 1,260 images.
We combine our result with DSS, and the performance is shown in the last row of Table~\ref{fig:average_precision}.
As can be seen, our model successfully wins in 10 out of 12 categories, and improves the mAP to $29.00\%$.
This improvement shows that our model can be applied to a variety of indoor scenes.
It is also extremely effective in improving the scene understanding result in the aligned sub-area.

\subsection{Room Layout and Total Scene Understanding}
\vspace{-1mm}
\paragraph{Layout estimation.}
As part of our model, we can estimate the existence and location of the ceiling, floor, and the wall directly behind the camera view. Table~\ref{tab:scene_confusion} shows quantitative evaluation. We can see that the 3D context network can successfully reduce the error and predict a more accurate room layout. Note that for some scene categories, the ceiling and wall are usually not visible from the images. These cases are marked as ``-''.
%For those situations, we put ``-'' in the table.

\begin{table}[t]
\center
\scalebox{0.80}{

\begin{tabular}{l|cccc}
\hline 
Layout Estimation & Sleeping & Office & Lounging & Table\tabularnewline
(Mean/Median) & Area & Area & Area & \&Chair\tabularnewline
\hline 
Ceiling Initial & 0.57/0.56 & - & - & 0.84/0.71\tabularnewline
Ceiling Estimate & 0.45/0.40 & - & - & 0.72/0.44\tabularnewline
Floor Initial & 0.30/0.25 & 0.28/0.24 & 0.25/0.23 & 0.22/0.20\tabularnewline
Floor Estimate & 0.10/0.09 & 0.09/0.06 & 0.22/0.16 & 0.08/0.05\tabularnewline
Wall Initial & 0.40/0.30 & 0.70/0.60 & - & -\tabularnewline
Wall Estimate & 0.22/0.08 & 0.60/0.21 & - & -\tabularnewline
\hline 
\end{tabular}
}
\vspace{1mm}
\caption{{\bf Error (in meter) for room layout estimation.} 
Our network reduces the layout error upon initialization from the transformation network. 
Note that for some scene categories, the ceiling and wall may not be visible from the images and therefore there are no annotations (marked with ``-'').
}
\label{tab:scene_confusion}
\vspace{-3mm}
\end{table}

\vspace{-3mm}
\paragraph{Scene understanding.}
We use the metrics proposed in \cite{SUNRGBD} to evaluate total 3D Scene Understanding accuracy. 
These metrics favor algorithms producing correct detections for all categories and accurate estimation of the free space.
We compare our model with Ren \etal (COG) \cite{Brown2016}.
% on the intersection of our testing set and their testing set.
%The authors of \cite{Brown2016} kindly provided their results to us. 
For geometry precision ($P_g$), geometry recall ($R_g$), and semantic recall ($R_r$), we achieve $71.02\%$, $54.43\%$, and $52.96\%$, which all clearly outperform $66.93\%$, $50.59\%$, and $47.99\%$ from COG.
Note that our algorithm uses only the depth map as input, while COG uses both color and depth.
%Also note that these two models use different training sets.
%COG uses more real RGB-D images than us, while we use our graphics synthesis algorithm to generate hybrid data for pretraining.

\begin{table}[t]
\center
\scalebox{0.85}{
\begin{tabular}{c|c|cccc}
\hline 
\multirow{2}{*}{Method} & \multirow{2}{*}{Sym.} & Sleeping & Office & Lounging  & Table\tabularnewline
 &  & Area & Area & Area & \&Chair\tabularnewline
\hline 
ICP & No & 75.6\% & 69.2\% & 58.5\% & 38.1\%\tabularnewline
ICP & Yes & 96.3\% & 89.0\% & 92.5\% & 75.3\%\tabularnewline
Network & No & 92.7\% & 87.9\% & 71.7\% & 44.3\%\tabularnewline
Network & Yes & 100.0\% & 93.4\% & 94.3\% & 73.2\%\tabularnewline
\hline 
\multicolumn{6}{c}{(a) Rotation Estimation Accuracy$\uparrow$}\tabularnewline
\multicolumn{6}{c}{}\tabularnewline
\hline 
\multirow{2}{*}{Method} & \multirow{2}{*}{Rot.} & Sleeping & Office & Lounging  & Table\tabularnewline
 &  & Area & Area & Area & \&Chair\tabularnewline
\hline 
ICP & - & 0.473 & 0.627 & 1.019 & 0.558\tabularnewline
Network & GT & 0.278 & 0.246 & 0.336 & 0.346\tabularnewline
Network & Est & 0.306 & 0.278 & 0.606 & 0.332\tabularnewline
\hline 
\multicolumn{6}{c}{(b) Translation Error (in meters) $\downarrow$}\tabularnewline
\end{tabular}
}
\caption{{\bf Evaluation of the transformation networks.} 
Our transformation network outperforms direct point cloud matching in the accuracies of both rotation and translation.
% dPlease refer to Section~\ref{sec:transformation_alignment} for details.
}
\label{fig:alignment_accuracy}
\vspace{-3mm}
\end{table}

% \vspace{-2mm}
\subsection{System Component Analysis}
\vspace{-2mm}
Our 3D context network relies on the initial alignment produced by scene template classification and transformation estimation model.
We also investigate how these factors affect our performance.

\vspace{-4mm}
\paragraph{Transformation Prediction.}
Table \ref{fig:alignment_accuracy} reports the evaluation of template alignment. For rotation, we show the percentage of data within a 10 degree range to the ground truth. For translation, we show the distance between the estimated translation and the ground truth.

For rotation, since some scenes (especially for lounging area and table\&chair set) are symmetric with respect to the horizontal plane, a correct estimation of the main direction would be enough for our purposes. 
Therefore, we report the accuracies both with and without symmetry [Sym.].

To compare with our neural network-based approach,
we design an ICP approach based on point cloud alignment as a baseline. 
Given a point cloud from a testing depth map,
we align it with the point cloud of each image in the training set, 
by exhaustively searching for the best rotation and translation,
using the measurement in Section~\ref{sec:hybrid}.
We choose the alignment with the best aligned training depth map as our transformation. 
We can see that our neural network based approach significantly outperforms this baseline.

To see how sensitive our model is to the initial alignment, we evaluate our model with the ground truth alignment, and the result is shown in Table~\ref{fig:average_precision} [GT Align].
We can see that the 9 out of 12 categories are improved in terms of AP, compared to that with estimated transformation, and the overall mAP improves $2.19\%$.

\vspace{-3mm}
\paragraph{Template Classification.}
The accuracy of the scene template classification is 89.5\%.
In addition to the ground truth transformation, we test our model with truth template category.
This further improves the mAP by $1.52\%$. %, which can be considered as an upper-bound performance of our system.

\vspace{-1mm}
\section{Conclusion}
\vspace{-1mm}
We propose a 3D ConvNet architecture that directly encodes context and local evidence leveraging scene template. 
The template is learned from training data to represent the functional area with relatively strong context evidence.
We show that context model provides complementary information with a local object detector, which can be easily integrate.
Our system has a fairly high coverage on real datasets, and achieves the state of the art performance for 3D object detection on the SUN-RGBD dataset. 
%There are a number of ways in which our work can be extended. The pipeline can be extended to use color information. With more annotated scenes, and we could define more scene templates to have a higher coverage of real images once more training images are provided. 
%At the system level, we plan to investigate ways to integrate the transformation alignment network and the 3D context network in a jointly differentiable manner \cite{jaderberg2015spatial} to enable end-to-end training.

\clearpage
{\small
\bibliographystyle{ieee}
\bibliography{panocontext,pvg,ref}
}

\end{document}